\documentclass[11pt,a4paper]{article}
\usepackage[hyperref]{acl2021}
\usepackage{times}
\usepackage{latexsym}
\usepackage{multirow}

\usepackage[utf8]{inputenc}
\usepackage{amsmath}

\usepackage{microtype}

\usepackage[normalem]{ulem}
\usepackage{breakurl}

\selectfont

\aclfinalcopy 

\title{Sockeye 3: Fast Neural Machine Translation with PyTorch}
\author{Felix Hieber\footnotemark[1], Michael Denkowski\footnotemark[1], Tobias Domhan\footnotemark[1], Barbara Darques Barros,\\\bf Celina Dong Ye, Xing Niu, Cuong Hoang, Ke Tran, Benjamin Hsu, Maria Nadejde,\\\bf Surafel Lakew, Prashant Mathur, Anna Currey, Marcello Federico\\
Amazon}

\date{}

\begin{document}

\maketitle

\renewcommand*{\thefootnote}{\fnsymbol{footnote}}
\footnotetext[1]{Corresponding authors:\newline \texttt{\{fhieber,mdenkows,domhant\}@amazon.com}}
\renewcommand*{\thefootnote}{\arabic{footnote}}

\begin{abstract}

Sockeye 3 is the latest version of the Sockeye toolkit for Neural Machine Translation (NMT). Now based on PyTorch, Sockeye 3 provides faster model implementations and more advanced features with a further streamlined codebase. This enables broader experimentation with faster iteration, efficient training of stronger and faster models, and the flexibility to move new ideas quickly from research to production. When running comparable models, Sockeye 3 is up to 126\% faster than other PyTorch implementations on GPUs and up to 292\% faster on CPUs. Sockeye 3 is open source software released under the Apache 2.0 license.

\end{abstract}

\section{Introduction}

Sockeye\footnote{\url{https://github.com/awslabs/sockeye}} provides a fast, reliable, and extensible codebase for Neural Machine Translation (NMT). As of version 3, Sockeye is based on PyTorch\footnote{\url{https://pytorch.org/}} \cite{NEURIPS2019_9015}, offering researchers a familiar starting point for implementing their ideas and running experiments. Sockeye's distributed mixed-precision training and quantized inference also enable users to quickly build production-ready NMT systems. Inference benchmarks show that Sockeye is up to 126\% faster than other PyTorch implementations on GPUs and up to 292\% faster on CPUs. Sockeye supports a range of advanced NMT features including source and target factors, source and target prefixes, lexical shortlists, and fast hybrid decoders. Sockeye powers Amazon Translate\footnote{\url{https://aws.amazon.com/translate/}} and has been used in numerous scientific publications.\footnote{\url{https://github.com/awslabs/sockeye\#research-with-sockeye}} Sockeye is developed as open source software on GitHub, where community contributions are welcome.

In the following sections, we describe Sockeye 3's scalable training (\S\ref{sec:training}), optimized inference (\S\ref{sec:inference}), and advanced features (\S\ref{sec:advanced}). We then share the results of a PyTorch NMT benchmark (\S\ref{sec:benchmark}) and case studies that use Sockeye features to implement formality and verbosity customization (\S\ref{sec:casestudy}). We conclude with a discussion of Sockeye's development philosophy (\S\ref{sec:development}) and include a minimal usage example in an appendix (\S\ref{sec:installation}).

\section{Training}
\label{sec:training}

Sockeye 3 implements optimized mixed precision training that scales to any number of GPUs and any size of training data.

\subsection{Parallel Data Preparation}

Sockeye provides an optional preprocessing step that splits training data into random shards, converts the shards to a binary format, and writes them to disk. During training, the shards are sequentially loaded and unloaded to enable training on arbitrarily large data with a fixed memory budget. Sockeye 3's data preparation step supports datasets of any size (subject to disk space) and runs in parallel on any number of CPUs. See Section~\ref{sec:sample} for instructions on how to run data preparation.

\subsection{Distributed Mixed Precision Training}
\label{sec:dist-training}
By default, Sockeye training runs in FP32 on a single GPU.  Activating mixed precision mode runs some or all of the model in FP16.\footnote{PyTorch AMP runs a mix of FP16 and FP32 operations to balance speed and precision: \url{https://pytorch.org/docs/stable/amp.html}. Apex AMP (O2) runs the entire model in FP16 to maximize speed: \url{https://nvidia.github.io/apex/amp.html}.}  This yields a direct speedup from faster calculations and an indirect speedup from fitting larger batches into memory.  Turning on distributed mode enables scaling to any number of GPUs by launching separate training processes that use PyTorch's distributed data parallelism\footnote{\url{https://pytorch.org/docs/stable/notes/ddp.html}} to synchronize updates.  In all cases, Sockeye traces the full encoder-decoder model with PyTorch's optimizing JIT compiler.\footnote{\url{https://pytorch.org/docs/stable/jit.html}}  Shown in Table~\ref{tab:training_gpus_precision}, activating mixed precision yields over 3X training throughput.  Scaling to 8 local GPUs yields 7.7X throughput, demonstrating 96.6\% GPU efficiency.

\begin{table}
    \centering
    \begin{tabular}{clr}
        GPUs & Precision & Tokens/Sec \\
        \hline
        1 & FP32 & 8,451 \\
        1 & FP16 \& FP32 & 28,287 \\
        8 & FP32 & 65,280 \\
        8 & FP16 \& FP32 & 218,688 \\
        \hline         
    \end{tabular}
    \caption{WMT14 En-De big transformer training benchmark on a p3.16xlarge EC2 instance using the large batch recipe described by \newcite{ott-etal-2018-scaling}.}
    \label{tab:training_gpus_precision}
\end{table}

\section{Inference}
\label{sec:inference}

Inference benefits from previous development for static computation graphs, avoiding dynamic shapes and data-dependent control flow as much as possible. As such, we are able to trace various components of the model with PyTorch's JIT compiler (encoder, decoder, and beam search).

\subsection{Quantization}

By default, Sockeye runs inference with FP32 model weights. Quantizing these weights to FP16 or INT8 can increase translation speed and reduce the model's memory footprint. Enabling FP16 quantization for GPUs casts the entire model to FP16 at runtime. Enabling INT8 quantization for CPUs activates PyTorch's dynamic quantization\footnote{\url{https://pytorch.org/tutorials/recipes/recipes/dynamic_quantization.html}} that runs linear layers (feed-forward networks) in INT8 while keeping the rest of the model in FP32. Both quantization strategies typically have minimal impact on quality and are recommended for most translation scenarios.

\subsection{Efficient Greedy Search}

\begin{table}
    \centering
    \begin{tabular}{lcc}
         \multicolumn{3}{r}{Translation Speed (Sent/Sec) $\uparrow$} \\
         & Baseline & Greedy \\
         \hline
         GPU FP16 & 11.9 & 13.1 \\
         +Lexical Shortlist & 12.0 & 13.9 \\
         \hline
         CPU FP32 & 2.6 & 2.7 \\
         +Quantize INT8 & 4.5 & 4.9 \\
         +Lexical Shortlist & 7.2 & 7.6 \\
         \hline
    \end{tabular}
    \caption{Benchmark comparing Sockeye's beam search with size 1 (baseline) to greedy search for a big transformer (WMT17 En-De) with batch size 1. GPU inference runs on a g4dn.xlarge EC2 instance and CPU inference runs on a c5.2xlarge EC2 instance.}
    \label{tab:inference_greedy}
\end{table}

Work on high performance NMT reports that certain types of models produce adequate translations without beam search \cite{junczys-dowmunt-etal-2018-marian-cost}. For such cases, Sockeye provides a dedicated implementation of greedy search that does not have the computational overhead of maintaining hypotheses in a beam. Table \ref{tab:inference_greedy} compares Sockeye's beam search with a beam size of 1 to the greedy implementation. Greedy search improves translation speed by 16\% on GPUs and 6\% on CPUs for a model that is already optimized for speed (quantized weights and lexical shortlists).

\begin{table*}
    \centering
    \begin{tabular}{llcccc}
        & & \multicolumn{2}{c}{big 6:6 transformer} & \multicolumn{2}{c}{big 20:2 transformer+SSRU} \\
        & & newstest & newstest-UP & newstest & newstest-UP \\
        \hline
        & baseline & 35.64 & 25.94 & 34.48 & 24.97 \\
        En-De & +SF & 35.52 & 28.85 & 34.62 & 27.37 \\
        & +SF+TF & 35.18 & \bf 33.47 & 35.12 & \bf 32.82 \\
        \hline
        & baseline & 32.99 & 24.47 & 33.53 & 25.99 \\
        Ru-En & +SF & 32.82 & 25.87 & 33.16 & 26.72 \\
        & +SF+TF & 33.18 & \bf 31.39 & 33.60 & \bf 31.49 \\
        \hline
    \end{tabular}
    \caption{BLEU scores of different models on newstest and its all-uppercased version (newstest-UP). Using target case factors (+SF+TF) achieves significantly higher BLEU than using source case factors alone (+SF) and the baseline for translating all-uppercased inputs. The training data is augmented with 1\% all-uppercased pairs.}
    \label{tab:factors}
\end{table*}

\section{Advanced Features}
\label{sec:advanced}

Sockeye 3 migrates Sockeye 2's advanced NMT features \cite{domhan-etal-2020-sockeye} from MXNet \cite{mxnet2015} to PyTorch. Sockeye 3 also introduces new features that are exclusive to the PyTorch version.

\subsection{Migrated Sockeye 2 Features}
\label{sec:migrated}

\noindent\textbf{Source Factors} \cite{sennrich-haddow-2016-linguistic}: Combine additional representations (factors) with source word embeddings before running the first encoder layer.  Factors can encode any pre-computable token level information such as original case or BPE type.  This approach enables combining the advantages of a smaller normalized vocabulary (more examples of each type in the training data) and a larger fine-grained vocabulary (distinguish between types with the same normalized form but different original forms).
\\

\noindent\textbf{Lexical Shortlists} \cite{devlin-2017-sharp}: When translating an input sequence, limit the target vocabulary to the top $k$ context free translations of each source token.  This can significantly increase translation speed by reducing the size of the output softmax that runs at each decoding step.  Sockeye provides tools for generating shortlists from the training data with \texttt{fast\_align}\footnote{\url{https://github.com/clab/fast_align}} \cite{dyer-etal-2013-simple} and uses a default value of $k=200$.

\subsection{SSRU Decoder}
\label{sec:ssru}

Sockeye supports replacing self-attention layers in the decoder with Simpler Simple Recurrent Units (SSRUs), which are shown to substantially improve translation throughput \citep{kim-etal-2019-research}. An SSRU simplifies the LSTM cell by removing the reset gate and replacing the $\tanh$ non-linearity with $\mathrm{ReLU}$:
\begin{align*}
\mathbf{f}_t &= \sigma(\mathbf{W}_t\mathbf{x}_t + \mathbf{b}_f) \\
\mathbf{c}_t &= \mathbf{f}_t \odot \mathbf{c}_{t-1} + (1-\mathbf{f}_t )  \odot \mathbf{W}\mathbf{x}_t  \\ 
\mathbf{h}_t &= \mathrm{ReLU}(\mathbf{c}_t)
\end{align*}
Only the cell state $\mathbf{c}_t$ requires sequential computations while other parts of the SSRU can be computed in parallel.

\subsection{Target Factors}

Factored models have been used to enrich phrase-based MT and NMT with linguistic features \citep{koehn-hoang-2007-factored,garcia-martinez-etal-2016-factored}. They reduce the output space by decomposing surface words $y$ on different dimensions, such as lemma and morphological tags, and maximize $\textrm{P}(y^t|y^{<t}\mathbf{x}) = \prod_{i=1}^n \textrm{P}(f_i^t|y^{<t},\mathbf{x})$.

When target factors are enabled, Sockeye 3 predicts target words ($f_1$) and attributes ($f_{2\dots n}$) with independent output layers, and the embeddings of the word and attributes are combined for the next decoder step. It incorporates the dependency between words and attributes by time-shifting attributes so that attributes at time $t$ are actually predicted at time $t+1$.

Following \citet{niu-etal-2021-faithful}, we test the effectiveness of using target case factors in translating all-uppercased inputs. We use the same train/dev/test data processing procedures as in other sections, except we (1) uppercase 1\% training pairs and add them back to the training; (2) truecase data and deduct case factors as detailed in \citet{niu-etal-2021-faithful}, and (3) additionally evaluate on all-uppercased newstest sets.

Results in Table~\ref{tab:factors} show that, with sub-optimal data augmentation, using target case factors (+SF+TF) achieves significantly higher BLEU scores than using source case factors alone (+SF) and the baseline for translating all-uppercased inputs.

\subsection{Fine-Tuning with Parameter Freezing}

When fine-tuning models, freezing some or most of the parameters can increase training throughput, avoid overfitting on small in-domain data, and yield compact parameter sets for multitask or multilingual systems \cite{wuebker-etal-2018-compact,JMLR:v22:20-1307}.  Sockeye supports freezing any model parameter by name as well as presets for freezing everything except a specific part of the model (decoder, output layer, embeddings, etc.).  When updating only the decoder, Sockeye turns off autograd for the encoder and skips its backward pass.  This yields faster training updates and lowers memory usage, which enables larger batch sizes.

\subsection{Source and Target Prefixes}
\label{sec:prefix}
Adding artificial source and target tokens has become a staple technique for NMT with applications ranging from multilingual models \cite{johnson-etal-2017-googles} to output length customization \cite{isometric_mt}. While these tokens can be added to training data with simple pre-processing scripts, adding them during inference requires extended support in the NMT toolkit. Sockeye 3 enables users to specify arbitrary prefixes (sequences of tokens) on both the source and target sides for any input. Source prefixes are automatically added to the beginning of each input. When inputs are split into multiple chunks,\footnote{During Sockeye inference, inputs that exceed the maximum sequence length set during training are split into smaller ``chunks'' that are translated independently.} the source prefix is included at the beginning of each chunk. When a target prefix is specified, Sockeye 3 forces the decoder to generate the $N$ prefix tokens as the first $N$ decoder steps before continuing the translation normally. Because target prefixes have diverse use cases, Sockeye 3 allows users to choose whether to apply prefixes when translating all input chunks and whether to strip prefixes out of the translation output. For instance, multilingual NMT requires special tokens to be added to each chunk to identify the output language, but removes these artificial tokens from the final translation. By contrast, continuing partial translations requires that the prefix is only added to the first chunk and includes that prefix as part of the translation.

As an example of leveraging special tokens, let us consider a multilingual NMT model where the output language is specified on the source side. Using this model to translate into German requires adding the token \texttt{<2DE>} to the beginning of each input. Such source prefixes can be specified using Sockeye's JSON input format:
\begin{verbatim}
{"text": "The boy ate the waff@@
le .", "source_prefix": "<2DE>"}
\end{verbatim}
This adds \texttt{<2DE>} to the beginning of each source chunk. If the model uses special target tokens to determine output language, a target prefix can be specified:
\begin{verbatim}
{"text": "The boy ate the waff@@
le .", "target_prefix": "<2DE>"}
\end{verbatim}
This forces the decoder to generate \texttt{<2DE>} as its first target token. Finally, Sockeye 3 supports adding source and target prefix factors. For example:
\begin{verbatim}
{"text": "The boy ate the waff@@
le .", "target_prefix": "<2DE>", 
"target_prefix_factors": ["O O B"
]}
\end{verbatim}
Here \texttt{<2DE>} is force-decoded as the first target token and aligns with target factor \texttt{O}. The next two target tokens after \texttt{<2DE>} are assigned target factors \texttt{O} and \texttt{B}. 

\subsection{Neural Vocabulary Selection}
\label{sec:nvs}
Instead of selecting the target vocabulary out of context as in lexical shortlisting (\S\ref{sec:advanced}), Neural Vocabulary Selection~(NVS) \cite{domhan-etal-2022} uses the encoder's hidden representation to predict the set of target words and is learned jointly with the translation model. Similarly, it results in lower translation latency via reduced computation per decoder step. The advantage of NVS lies in its simplicity, as no external alignment model is required and predictions are made in context, resulting in a higher recall of target words for given target vocabulary size. 

\section{Benchmark}
\label{sec:benchmark}

\begin{table*}
    \centering
    \begin{tabular}{lcccc}
         & \multicolumn{2}{c}{WMT17 En-De} & \multicolumn{2}{c}{WMT17 Ru-En} \\
         & Training Time (Hours) $\downarrow$ & BLEU $\uparrow$ & Training Time (Hours) $\downarrow$ & BLEU $\uparrow$ \\
         \hline
         Sockeye &  9.9 & 35.3 & 28.1 & 33.1 \\
         Fairseq & 10.0 & 35.3 & 28.0 & 33.0 \\
         OpenNMT & 13.7 & 35.2 & 39.4 & 32.2 \\
         \hline
    \end{tabular}
    \caption{Big transformer training benchmark using 8 GPUs on a p3.16xlarge EC2 instance.  Models are trained using the large batch recipe described by \newcite{ott-etal-2018-scaling} for either 25K (En-De) or 70K updates (Ru-En).}
    \label{tab:training}
\end{table*}

\begin{table*}
    \centering
    \begin{tabular}{lccc}
         & \multicolumn{3}{c}{Translation Speed (Sent/Sec) $\uparrow$} \\
         & Sockeye & Fairseq & OpenNMT \\
         \hline
         GPU FP16 Batch 64 & 67.8 & 66.1 & 47.8 \\
         +Lexical Shortlist & 76.0 & -- & -- \\
         \hline
         GPU FP16 Batch 1 & 8.4 & 3.2 & 4.2 \\
         +Lexical Shortlist & 9.5 & -- & -- \\
         \hline
         CPU FP32 Batch 1 & 1.2 & 1.1 & 1.2 \\
         +Quantize INT8 & 2.4 & -- & -- \\
         +Lexical Shortlist & 4.7 & -- & -- \\
         \hline
    \end{tabular}
    \caption{Big transformer WMT17 En-De inference benchmark. GPU inference runs on a g4dn.xlarge EC2 instance and CPU inference runs on a c5.2xlarge EC2 instance.  All reported values are averages over 3 runs.  The listed techniques do not significantly impact translation quality; BLEU scores for all settings are within 0.2 of the FP32 baseline.}
    \label{tab:inference}
\end{table*}

\begin{table*}
    \centering
    \begin{tabular}{lcccc}
         & \multicolumn{2}{c}{WMT17 En-De} & \multicolumn{2}{c}{WMT17 Ru-En} \\
         & Training Time (Hours) $\downarrow$ & BLEU $\uparrow$ & Training Time (Hours) $\downarrow$ & BLEU $\uparrow$ \\
         \hline
         Big 6:6       &  9.9 & 35.3 & 28.1 & 33.1 \\
         Big 20:2      & 14.7 & 34.7 & 41.2 & 33.5 \\
         Big 20:2 SSRU & 15.6 & 34.9 & 44.2 & 33.0 \\
         \hline
    \end{tabular}
    \caption{Sockeye model architecture training benchmark using 8 GPUs on a p3.16xlarge EC2 instance.  Models are trained using the large batch recipe described by \newcite{ott-etal-2018-scaling} for either 25K updates (En-De) or 70K updates (Ru-En).  Model checkpoints are saved every 500 updates and the 8 best checkpoints are averaged.}
    \label{tab:training_alternate}
\end{table*}

\begin{table*}
    \centering
    \begin{tabular}{lccc}
         & \multicolumn{3}{c}{Translation Speed (Sent/Sec) $\uparrow$} \\
         & Big 6:6 & Big 20:2 & Big 20:2 SSRU \\
         \hline
         GPU FP16 Batch 64 & 73.8 & 116.3 & 142.8 \\
         GPU FP16 Batch 1 & 9.9 & 17.4 & 18.5 \\
         CPU INT8 Batch 1 & 4.5 & 7.8 & 9.5 \\
         \hline
    \end{tabular}
    \caption{Sockeye model architecture WMT17 En-De inference benchmark. GPU inference runs on a g4dn.xlarge EC2 instance and CPU inference runs on a c5.2xlarge EC2 instance.  All configurations use lexical shortlists.  All reported values are averages over 3 runs.}
    \label{tab:inference_alternate}
\end{table*}

We conduct a reproducible benchmark of PyTorch-based neural machine translation toolkits that includes Sockeye, Fairseq\footnote{\url{https://github.com/pytorch/fairseq}} \cite{ott2019fairseq}, and OpenNMT\footnote{\url{https://github.com/OpenNMT/OpenNMT-py}} \cite{klein-etal-2017-opennmt}.  For each toolkit, we train a standard big transformer model and run inference on GPUs and CPUs.  The scripts used to conduct this benchmark are publicly available.\footnote{\url{https://github.com/awslabs/sockeye/tree/arxiv_sockeye3/arxiv}}

\subsection{Training}
\label{sec:benchmark_training}

We select two translation tasks for which pre-processed data sets are available: WMT17 English-German (5.9M sentences) and Russian-English (25M sentences).\footnote{\url{https://data.statmt.org/wmt17/translation-task/preprocessed}}  We further process the data by applying byte-pair encoding\footnote{\url{https://github.com/rsennrich/subword-nmt}} \cite{sennrich-etal-2016-neural} with 32K operations and filtering out sentences longer than 95 tokens.  We use each toolkit to train a big transformer model \cite{vaswani2017attention} on 8 local GPUs (p3.16xlarge EC2 instance) using the large batch recipe described by \newcite{ott-etal-2018-scaling}.  Models are trained for either 25K updates (En-De) or 70K updates (Ru-En) with checkpoints every 500 updates.  The 8 best checkpoints are averaged to produce the final model weights.

We use the fastest known settings for each toolkit that do not change the model architecture or training recipe. This includes enabling NVIDIA's Apex\footnote{\url{https://github.com/NVIDIA/apex}} extensions for PyTorch and running the entire model in FP16. Shown in Table~\ref{tab:training}, Sockeye and Fairseq are fastest, training models with comparable BLEU scores in comparable time. 

\subsection{Inference}

We benchmark inference on GPUs (g4dn.xlarge EC2 instance) and CPUs (c5.2xlarge EC2 instance with 4 physical cores).  Shown in Table~\ref{tab:inference}, Sockeye matches or outperforms other toolkits on GPUs and CPUs with and without batching.  When activating NMT optimizations that are only natively supported by Sockeye (lexical shortlists and CPU INT8 quantization\footnote{At the time of writing, activating OpenNMT's INT8 mode does not appear to have any impact.}), Sockeye is fastest across the board: +15\% for batched GPU inference, +126\% for non-batched GPU inference, and +292\% for CPU inference.

\subsection{Alternate Model Architectures}
\label{sec:alternate}

\newcite{domhan-etal-2020-sockeye} report that transformer models with deep encoders and shallow decoders (20:2) can translate significantly faster than standard models (6:6) with similar quality ($\pm$1 BLEU).  The speedup can be attributed to better parallelization in the encoder (sequence-level operations versus per-step operations) and fewer calculations per encoder layer (no encoder-decoder cross-attention and a single forward pass versus beam search).

We benchmark three versions of Sockeye's transformer: (1) the standard big 6:6 model from Section~\ref{sec:benchmark_training}, (2) a big 20:2 model, and (3) a big 20:2 model that replaces decoder self-attention with SSRUs as described in Section~\ref{sec:ssru}.  Shown in Tables \ref{tab:training_alternate} and \ref{tab:inference_alternate}, moving from a 6:6 model to a 20:2 model yields up to a 76\% inference speedup and moving to SSRUs yields up to a 23\% additional speedup (87\%-111\% faster than the baseline).  These models do take longer to train.  The 20:2 transformers have substantially more parameters (46 sub-layers versus 30) and SSRUs do not parallelize as well as self-attention during training.  These trade-offs make models with deep encoders, shallow decoders, and SSRUs a good match for tasks where decoding time and costs dominate.  This includes experiments that translate large amounts of data (e.g., back-translation) and applications where NMT models are deployed for large volume translation.

\section{Case Studies}
\label{sec:casestudy}
\subsection{Formality Control}
We present a case study on using Sockeye 3 transformer models with deep encoders and shallow SSRU decoders (20:2) introduced in Sections ~\ref{sec:ssru} and \ref{sec:alternate}, and the source prefix feature introduced in Section~\ref{sec:prefix}. We train unconstrained baseline and formality controlled models for 6 language pairs for the 2022 IWSLT shared task on Formality Control for Spoken Language Translation.\footnote{\url{https://iwslt.org/2022/formality}} The baseline models and fine-tuning instructions are publicly available.\footnote{ \url{https://github.com/amazon-research/contrastive-controlled-mt/tree/main/IWSLT2022/models}}

The English-German and English-Spanish models were trained on 20M pairs sampled from ParaCrawl v9 \cite{banon-etal-2020-paracrawl}, using WMT newstest for development. The English-Japanese model was trained on all 10M pairs from JParaCrawl v2 \cite{morishita-etal-2020-jparacrawl} using the IWSLT17 development set. 
The English-Hindi model was trained on all 15M pairs from CCMatrix \cite{schwenk-etal-2021-ccmatrix}, using the WMT newsdev2014 for development and newstest2014 for testing.

For evaluating generic quality, we used the WMT newstests\footnote{We used newstest 2020 for German, 2014 for Spanish, 2014 for Hindi, 2020 for Japanese} as well as the MuST-C test sets \cite{di-gangi-etal-2019-must}. To train and evaluate formality-controlled models we use the CoCoA-MT dataset and benchmark \cite{nadejde-etal-2022-coca-mt}. We replicate the experiments in \citeauthor{nadejde-etal-2022-coca-mt} using Sockeye 3 models: we  fine-tune the generic baseline MT model on labeled contrastive translation pairs augmented by an equal number of randomly sampled unlabeled generic training data. The contrastive translation pairs are labeled using a special source prefix that specifies the formality level of the target: 
\begin{verbatim}
src: <FORMAL> `Are you tired?`
trg: `Sind Sie muede?`
src: <INFORMAL> `Are you tired?`
trg: `Bist du muede?`
\end{verbatim}
At inference time, we use the source prefix to control the formality level in the output. We report evaluation results in Table~\ref{tab:AccuracyAllLPs} showing formality-controlled models have high targeted accuracy while preserving generic quality.

\begin{table*}
\centering
\small{
\begin{tabular}{l  l |ccc|cc}
\hline

&  & \multicolumn{3}{c|}{M-ACC - CoCoA-MT test}  & \multicolumn{2}{c}{BLEU } \\ \hline
Lang. & System &  F & I & Avg. & WMT & TED \\
\hline \hline
\multirow{2}{1cm}{\centering EN-DE} & generic & -  & -  & -  & 42.1 & 32.7   \\
 & controlled &  97.8  & 45.0  & 71.4  & 41.4 & 32.1   \\ \hline
\multirow{2}{1cm}{\centering EN-ES} & generic & -  & -  & -  & 35.1 & 36.7  \\
 & controlled &  89.1  & 47.8  & 68.4  & 35.0 & 36.9   \\ \hline
\multirow{2}{1cm}{\centering EN-HI} & generic & -  & -  & -  & 10.0 &  - \\ 
 & controlled &  96.3  & 36.7  & 66.5  & 9.9 & -  \\  \hline
\multirow{2}{1cm}{\centering EN-JA} & generic &  -  & -  & -  & 21.7 & 14.3  \\
 & controlled &  68.8  & 83.2  & 76.0  & 22.2 & 14.3   \\ 
 \hline 
\end{tabular}
}
\caption{Accuracy of generic baseline and formality-controlled models on the CoCoA-MT test set. The TED test sets are MuST-C for EN-DE,ES and IWSLT for EN-JA. For controlled models, M-Acc (F)/(I) scores are computed using formal/informal translations respectively, resulting in performance upper bounds of 100\%.  
\label{tab:AccuracyAllLPs} }
\end{table*}

\begin{table*}[!htb]
\small
\centering
\begin{tabular}{c|c|c|ccc}
Lang. Pair             & Test set                     & System    & BERTScore & LC    & BERTScore$\times$LC \\ \hline
\multirow{5}{*}{En-De} & \multirow{2}{*}{MuST-C}      & Baseline  & 0.837     & 41.3 & 34.6        \\
                       &                              & VC        & 0.834     & 56.6 & 47.2        \\ \cline{2-6}
                       & \multirow{3}{*}{IMT} & Baseline         & 0.757     & 51.5 & 39.0        \\
                       &                              & VC        & 0.757     & 56.5 & 42.8        \\ 
                       &                              & VC+Rank   & 0.743	  & 63.5 & 47.2       \\ \hline
\multirow{5}{*}{En-Fr} & \multirow{2}{*}{MuST-C}      & Baseline  & 0.867     & 38.7 & 33.6        \\
                       &                              & VC        & 0.860     & 53.6 & 46.1        \\ \cline{2-6}
                       & \multirow{3}{*}{IMT} & Baseline         & 0.778     & 39.5 & 30.7        \\
                       &                              & VC        & 0.778     & 58.0 & 45.1        \\ 
                       &                              & VC+Rank   & 0.772	  & 65.5 & 50.6      \\ \hline
\multirow{5}{*}{En-Es} & \multirow{2}{*}{MuST-C}      & Baseline  & 0.846     & 60.0 & 50.8        \\
                       &                              & VC        & 0.845     & 66.7 & 56.4        \\ \cline{2-6}
                       & \multirow{3}{*}{IMT} & Baseline         & 0.802     & 59.0 & 47.3        \\
                       &                              & VC        & 0.799     & 62.5 & 50.0        \\
                       &                              & VC+Rank   & 0.789	  & 64.0 & 50.5    \\ \hline
\end{tabular}
\caption{Results comparing a standard NMT model (Baseline), NMT with output verbosity control (VC), and VC with N-best re-ranking (VC+Rank) on the Ted Talks MuST-C test set released for the isometric MT (IMT) shared task. Models are evaluated using BERTScore and length compliance within $\pm$10\% (LC), and the final system ranking metric (BERTScore$\times$LC).} 
\label{tab:isometric-mt}
\end{table*}

\subsection{Isometric MT}
We present another case study on Isometric MT where the task is to generate translations similar in length to the source text. In this setup, we experiment with the verbosity control (VC) work of \citet{lakew-etal-2019-controlling}, specifically the length token approach using Sockeye's source prefix implementation (Section \ref{sec:prefix}). 
We train and evaluate models using data from the constrained setting in the 2022 IWSLT shared task on Isometric Spoken Language Translation.\footnote{\url{https://iwslt.org/2022/isometric}} 
Evaluation is done on MuST-C v1.2~\cite{cattoni2021mustc} and a blind set released as part of the shared task.\footnote{\url{https://github.com/amazon-research/isometric-slt/tree/main/dataset}}
All models are evaluated on three language pairs: English-German, English-French, and English-Spanish using BERTScore~\cite{bert-score}. There is also a length compliance (LC) metric~\cite{isometric_mt} which measures whether the translation is within $\pm10\%$ of the source length and a  final system ranking metric that combines BERTScore and LC.
For preprocessing, we leverage SentencePiece~\cite{kudo-richardson-2018-sentencepiece} with 16.5K operations.
Models are trained with the transformer base (6:6) architecture on 8 GPUs (p3.16xlarge instances).
At training time, we apply \texttt{<short>}, \texttt{<normal>}, and \texttt{<long>} prefixes to the source side of the parallel training data based on the length compliance of the target side. 
During inference, we add the \texttt{<normal>} prefix to generate translations that are similar in length to source. Following~\cite{lakew-etal-2019-controlling}, we also run an ablation study for the blind set where we re-rank the $N$-best list to find the best translation in terms of translation quality and length. We report results in Table~\ref{tab:isometric-mt} that show improvements when adding verbosity control (VC) to baseline models and further improvements when applying $N$-best re-ranking (VC+Rank).

\section{Development}
\label{sec:development}

Sockeye is developed as open source software under the Apache 2.0 license and hosted on GitHub. All contributions are publicly reviewed using GitHub's pull request system. Sockeye is written in PEP 8 compatible Python 3 code. Functions are documented with Sphinx-style docstrings and include type hints for static code analysis. Sockeye includes an extensive suite of unit, integration, and system tests covering the toolkit's core functionality and advanced features. New code is required to pass all tests (and add new tests to cover new functionality) plus type checking and linting in order to be merged. Sockeye 3 retires some older features such as lexical constraints. We welcome pull requests from community members interested in porting these features from Sockeye 2.

\section{Acknowledgements}

We would like to thank Vincent Nguyen for helping us configure OpenNMT for our benchmark.

\bibliographystyle{acl_natbib}
\bibliography{references}

\clearpage
\appendix

\section{Installation and Usage}
\label{sec:installation}

\subsection{Installation}
The easiest way to install Sockeye is via pip:
\begin{verbatim}
> pip3 install sockeye
\end{verbatim}
Once Sockeye is installed, you can use the included command-line tools to train models (\texttt{sockeye-train}), translate data (\texttt{sockeye-translate}), and more. 
If you plan to extend or modify the code, you can install Sockeye from source:
\begin{verbatim}
> git clone https://github.com/
awslabs/sockeye.git
> cd sockeye 
> pip3 install --editable ./
\end{verbatim}
Using the \texttt{editable} flag means that changes to the code will apply directly without needing to reinstall the package.

\subsection{Sample Usage}
\label{sec:sample}
Training a Sockeye model requires parallel (source and target) training and validation data.
You can use raw training data directly, though we recommend using \texttt{sockeye-prepare-data} to prepare the data ahead of time.
This reduces memory consumption and data loading time during training:
\begin{verbatim}
> sockeye-prepare-data \
  -s [source training data] \
  -t [target training data] \
  -o [output directory]
\end{verbatim}

To train a model from scratch, run \texttt{sockeye-train} with the prepared training data directory, validation source and target data files, model output directory, and at least one stopping criteria such as number of training steps:
\begin{verbatim}
> sockeye-train \
  -d [prepared training data] \
  -vs [source validation data] \
  -vt [target validation data] \
  -o [output directory] \
  --max-updates [training steps]
\end{verbatim}
To fine-tune an existing model on new data (e.g., for domain adaptation), run \texttt{sockeye-train} with the new data and specify a checkpoint from the existing model with the \texttt{--params} argument.

Once you have trained a Sockeye model, you can use it to translate inputs by running:
\begin{verbatim}
> sockeye-translate \
 -m [model directory]
\end{verbatim}

This section covers a minimal example of using Sockeye's CLI tools.  For a step-by-step tutorial on training a standard transformer model on any size of data, see the WMT 2014 English-German example\footnote{\url{https://github.com/awslabs/sockeye/blob/main/docs/tutorials/wmt_large.md}} on GitHub.

\end{document}